\documentclass[letterpaper, 10 pt, conference]{ieeeconf}  %

\IEEEoverridecommandlockouts                              %

\usepackage{amsmath,amsfonts}
\usepackage{algorithm}
\usepackage{algpseudocode}
\usepackage{array}
\usepackage[caption=false,font=normalsize,labelfont=sf,textfont=sf]{subfig}
\usepackage{textcomp}
\usepackage{stfloats}
\usepackage{url}
\usepackage{verbatim}
\usepackage{graphicx}
\usepackage{cite}
\usepackage{bm}
\usepackage{xspace}
\usepackage{multirow}
\usepackage{booktabs}
\usepackage{rotating}
\usepackage{braket}
\usepackage{caption}
\usepackage{graphicx}
\usepackage{caption}
\usepackage{cuted}%
\usepackage{afterpage}%
\usepackage{soul}
\usepackage[dvipsnames]{xcolor}
\usepackage{mathtools}
\usepackage{todonotes}
\usepackage[export]{adjustbox}
\usepackage{xcolor,colortbl}
\usepackage{siunitx}
\makeatletter
\DeclareRobustCommand\onedot{\futurelet\@let@token\@onedot}
\def\@onedot{\ifx\@let@token.\else.\null\fi\xspace}
\def\eg{\emph{e.g}\onedot} 
\def\ie{\emph{i.e}\onedot} 
 
\def\etc{\emph{etc}\onedot} 
\def\wrt{w.r.t\onedot} 
 
\def\etal{\emph{et al}\onedot}

\definecolor{lavender}{rgb}{0.9, 0.9, 0.98}
\usepackage{soul}
\renewcommand\st[1]{\unskip}
\title{\LARGE \bf
Spatiotemporal Multi-Camera Calibration using Freely Moving People}

\author{Sang-Eun Lee$^{1}$, Ko Nishino$^{1}$, and Shohei Nobuhara$^{2}$
\thanks{$^{1}$ Graduate School of Informatics, Kyoto University, Kyoto 606-8501, Japan (e-mail: slee@vision.ist.i.kyoto-u.ac.jp; nishino.ko.5a@kyoto-u.ac.jp).
}
\thanks{$^{2}$ Information and Human Sciences, Kyoto Institute of Technology, Kyoto, 606-8585, Japan (e-mail: nob@kit.ac.jp).}
\thanks{©2025 IEEE. Personal use of this material is permitted. Permission from IEEE
must be obtained for all other uses, in any current or future media, including
reprinting/republishing this material for advertising or promotional purposes,
creating new collective works, for resale or redistribution to servers or lists, or
reuse of any copyrighted component of this work in other works.}
}
\begin{document}

\maketitle

\begin{strip}
\begin{minipage}{\textwidth}\centering
\vspace{-70pt}

\begin{center}
    \centering
    \captionsetup{type=figure}
    \begin{tabular}{ccc}
  \includegraphics[width=.32\linewidth,valign=m]{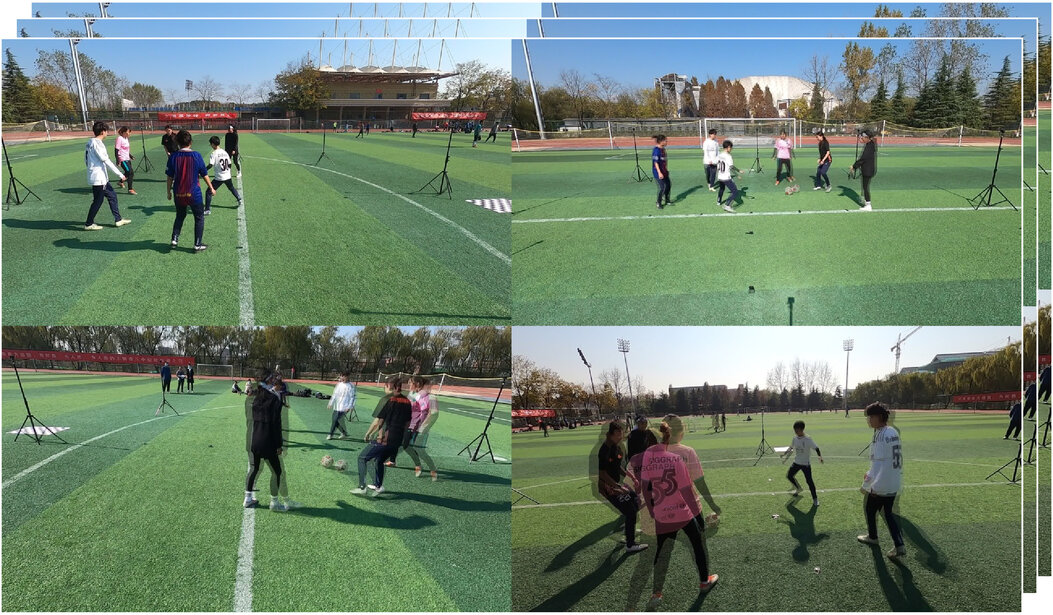} &
  \includegraphics[width=.32\linewidth,valign=m]{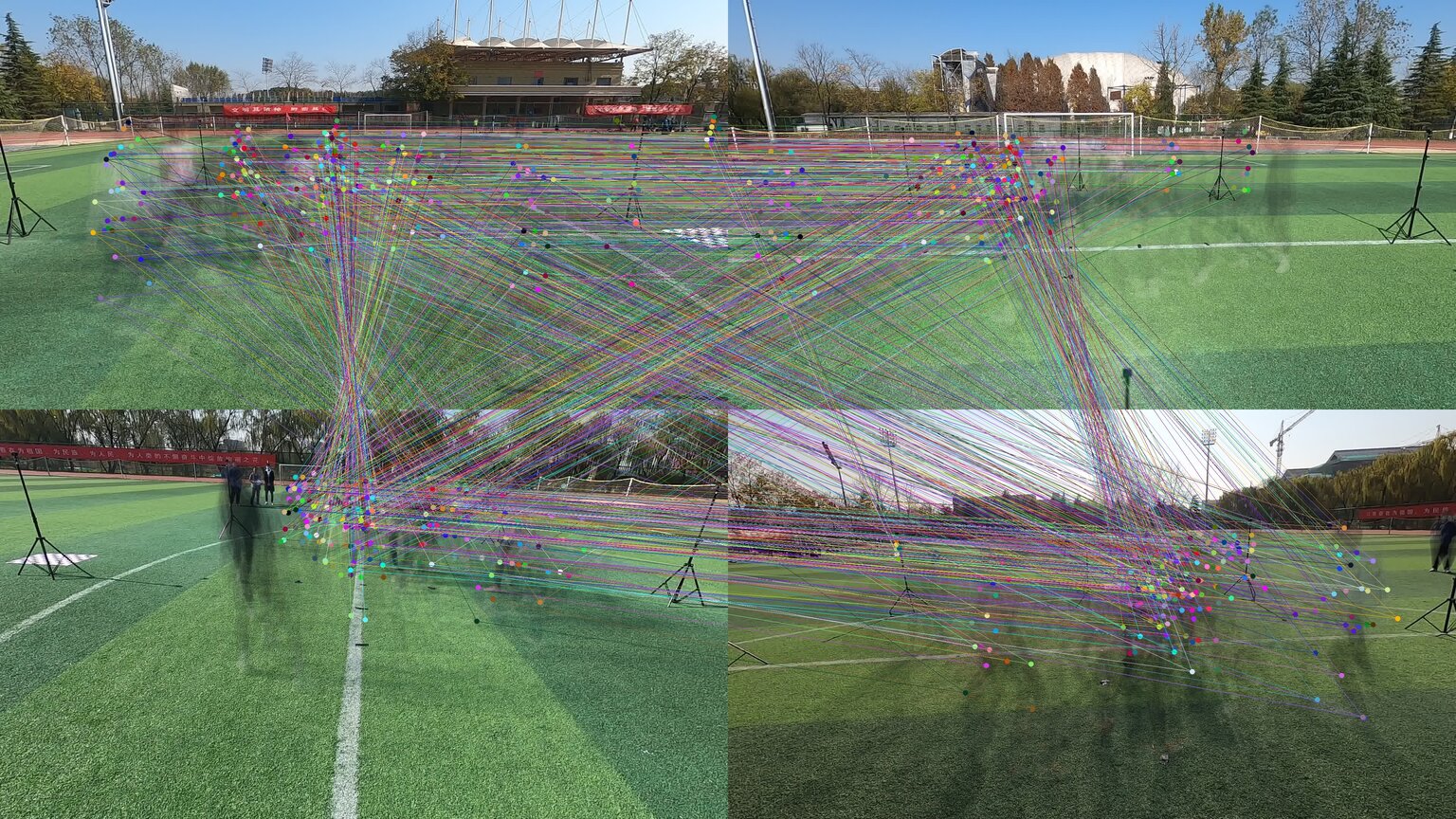} &
  \includegraphics[width=.24\linewidth,valign=m]{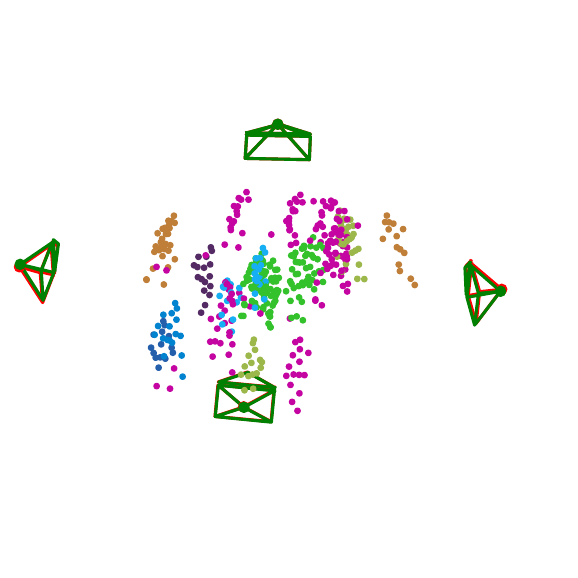} \\[-13pt]
    \multicolumn{1}{c}{ Unsync. \& Uncalib. Multi-view Videos} &
   \multicolumn{1}{c}{Spatiotemporal  Calibration \& Association  } &
   \multicolumn{1}{c}{Scene Reconstruction } \\
  \end{tabular}
  \vspace*{-1mm}
    \captionof{figure}{Our method estimates camera extrinsics, relative temporal offsets, and person associations directly from multi-view videos of freely moving people. This approach eliminates the need for specialized tools and burdensome procedures of calibration and synchronization procedures. As a result, we achieve spatio-temporal multi-camera calibration with reliably established correspondences, facilitating the accurate reconstruction of scenes involving multiple people and cameras.}
\label{fig:figure1}
\end{center}%

\end{minipage}
\vspace{-13pt}
\end{strip}

\begin{abstract}

We propose a novel method for spatiotemporal multi-camera calibration using freely moving people in multi-view videos. Since calibrating multiple cameras and finding matches across their views are inherently interdependent, performing both in a unified framework poses a significant challenge. We address these issues as a single registration problem of matching two sets of 3D points, leveraging human motion in dynamic multi-person scenes. To this end, we utilize 3D human poses obtained from an off-the-shelf monocular 3D human pose estimator and transform them into 3D points on a unit sphere, to solve the rotation, time offset, and the association alternatingly. We employ a probabilistic approach that can jointly solve both problems of aligning spatiotemporal data and establishing correspondences through soft assignment between two views. The translation is determined by applying coplanarity constraints. The pairwise registration results are integrated into a multiview setup, and then a nonlinear optimization method is used to improve the accuracy of the camera poses, temporal offsets, and multi-person associations. Extensive experiments on synthetic and real data demonstrate the effectiveness and flexibility of the proposed method as a practical marker-free calibration tool.

\end{abstract}

\section{INTRODUCTION}
Multi-view camera system plays an important role in 3D reconstruction, motion capture, surveillance, robotics, \etc.  These systems require the cameras to be calibrated and synchronized beforehand.  This point, however, makes it burdensome to deploy a multi-camera system since the existing calibration and synchronization methods require special devices and careful \textcolor{black}{on-site} operations~\cite{zhang2000flexible,aruco2014,hasler2009markerless}.

Can we automate this process without using additional devices and procedures?  For example, if we can calibrate and synchronize cameras from videos of a single event of interest (\eg a football match) taken by different people from different unknown positions themselves, we can turn such accidental multi-view videos into a 3D motion analysis system (Fig.~\ref{fig:figure1}).  Toward this goal, this paper proposes a new spatiotemporal multi-camera calibration that automatically finds corresponding points between views for spatial calibration while estimating their relative temporal offsets. \textcolor{black}{This makes multi-view vision systems more flexible and widely applicable.}

To allow calibration without additional devices, we utilize freely-moving persons in the videos themselves~\cite{leeRAL2022,yan2021wide-baseline,elhayek2012feature,nakano2021}.  Our method automatically associates persons in different views, and uses their joints as corresponding points for spatiotemporal calibration.  Here, the association and the calibration are fundamentally interdependent. The calibration requires to establish correspondences across views, while the association exploits the calibration that defines the epipolar constraints guides matching.

\begin{figure*}[tp]
\begin{center}

\includegraphics[width=1.\linewidth]{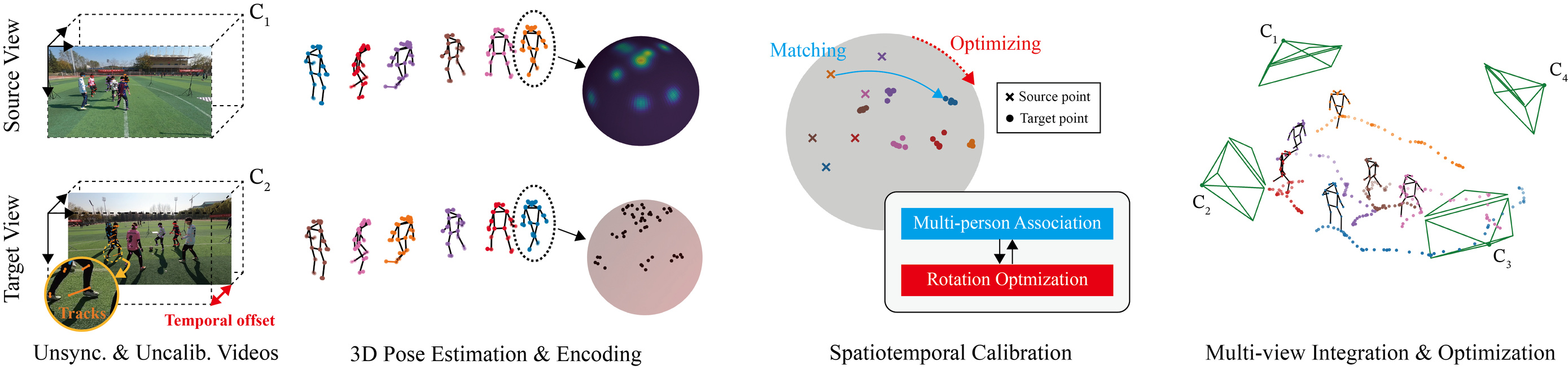}
 \end{center}
  \vspace*{-4mm}
   \caption{
	Pipeline of our method. For each input video, we perform 3D pose estimation and tracking from off-the-shelf models. The 3D poses are encoded as a set of 3D points on a unit sphere. For the source view, these directional data are modeled with directional statistical distribution. We align the 3D points from the target view with this parameterized model through an alternating optimization strategy. The calibration and association results are integrated into a multi-view system, which is then optimized to ensure temporal and geometric consistency using non-linear optimization.
	}
 
\label{fig:overview}
\end{figure*}

Our key idea is to solve these two problems, \ie, calibration and association, as a single registration problem of two sets of 3D points on a unit sphere (Fig. \ref{fig:overview}).  Given a pair of uncalibrated videos without association and synchronization, our method finds persons in each view, estimates their 3D poses by utilizing an off-the-shelf monocular 3D pose estimator, and encodes the 3D poses as a set of points on a unit sphere. By finding a 3D rotation that aligns these 3D points, our method can determine the relative rotation of the cameras together with person associations simultaneously. We can also incorporate temporal alignment between the videos within this framework. Our method then utilizes an existing method to determine translation using the estimated rotation and associations, with known intrinsic parameters given in advance~\cite{Hartley00}. After applying Pose Graph Optimization (PGO)~\cite{kummerle2011g} to ensure global consistency of the estimates of paired views in a multi-view setup, we further enhance the accuracy of our method using non-linear optimization. 

We show that our method is accurate as well as robust to occlusions, through an extensive set of experiments on both synthetic and real datasets. We also compare it with the state-of-the-art approaches, including point-set registration problem~\cite{Peng2022ARCSAR}, extrinsic camera calibration using people in a multi-view system~\cite{yan2021wide-baseline}, and spatiotemporal bundle adjustment leveraging the motion of moving people~\cite{Vo2020SpatiotemporalBA}.  These results demonstrate that the proposed method achieves a practical marker-free calibration and synchronization in real-world scenarios.

\section{Related Work}

\subsection{Point-set Registration} 

Point-set registration estimates a transformation matrix to align two point clouds. Given correspondences between the points, Procrustes analysis~\cite{gower1975generalized} finds the similarity transform matrix while orthogonal Procrustes analysis~\cite{Schnemann1966AGS} finds an orthogonal matrix. Iterative Closest Point (ICP)~\cite{besl1992method} finds a rigid transformation and correspondences simultaneously.

Many variants of ICP~\cite{myronenko2010point,segal2009generalized} have been proposed to address limitations, such as initialization, input noise, and outliers.
Probabilistic approaches~\cite{myronenko2010point,Parkison2018SemanticIC} efficiently manage uncertainties in establishing correspondences. Semantic ICP~\cite{Parkison2018SemanticIC} integrates semantic cues to enhance the reliability of the registration against noise and outliers. Several investigations~\cite{Peng2022ARCSAR,yang2019quaternion} focus on the rotation search algorithm rather than the full rigid transformation for minimal inlier data. In particular, ARCS~\cite{Peng2022ARCSAR} achieves robust and efficient performance in simultaneous rotation and correspondence search on large-scale datasets.

Compared with these methods, our approach translates camera calibration into 3D \textit{oriented} point sets alignment in a probabilistic framework.
This allows calibrating cameras with multiple instances (\eg persons) lookalike each other in their appearances.

\subsection{Spatiotemporal Multi-camera Calibration}

\subsubsection{Multi-camera Calibration} 

A common approach for multi-camera calibration utilizes reference objects such as checkerboard pattern~\cite{Zhang96} and ArUco markers~\cite{aruco2014} to facilitate establishing correspondences across multiple images. These methods, however, can be cumbersome and require careful operations. Leveraging dynamic foreground objects in our daily lives, such as freely moving people in scenes, has been proposed as a potential alternative to such manual calibration~\cite{Takahashi18, leeRAL2022,Lv_pami2006_walking,nakano2021,albl2017two}. Using human heights in motion, represented as vertical line segments perpendicular to the ground plane, has demonstrated the possibility for calibration~\cite{Lv_pami2006_walking,nakano2021}. This approach facilitated the estimation of vanishing points~\cite{Lv_pami2006_walking} and provided closed-form solutions for the line segments~\cite{nakano2021}. Additionally, human keypoints obtained from 2D human pose estimation have served as the calibration pattern~\cite{Takahashi18}. Moreover, 3D human poses, derived from elevating 2D human poses, have been employed as reference targets for calibration~\cite{leeRAL2022,Garau2020}. Lee~\etal~\cite{leeRAL2022} introduced a linear calibration algorithm based on factorization using 3D oriented points extracted from 3D human poses.

The methods discussed so far assume that the correspondences between views are determined beforehand, which lacks the versatility required for dynamic, multi-person scenes. To overcome this challenge, various priors, such as geometry, appearance, and motion, have been explored~\cite{huang2021dynamic,yan2021wide-baseline,xu2023multi,li2024multi}. Huang~\etal~\cite{huang2021dynamic} jointly recovers multiple human motions and extrinsic camera parameters using physics-based constraints to denoise 2D poses of pre-established correspondences and ensure temporal coherence with a human motion prior. Yan~\etal~\cite{yan2021wide-baseline} employed a re-identification (re-ID) network to associate bounding boxes across different camera views, then converted these into point correspondences to calculate relative camera poses across camera pairs. Xu~\etal~\cite{xu2023multi} treated the cross-view matching problem as a clustering problem using 2D appearance features within uncalibrated camera networks. Li~\etal\cite{li2024multi} uses 3D Re-ID appearance features from RGBD images to establish correspondences, which facilitates depth-based camera pose and 3D multi-human pose estimation. 

\subsubsection{Multi-camera Synchronization}

The most straightforward way for aligning timing differences between image sequences is to use wired connections with hardware triggers in a camera network~\cite{Sigal2010HumanEvaSV}. Alternatively, audio triggering, which employs microphones to detect sound signals or audio cues, offers another mean of synchronization, particularly useful in setups where a wired connection may not be feasible~\cite{Joo_2017_TPAMI,hasler2009markerless}. However, this often requires laborious manual intervention conducted by experts in controlled environments. In contrast, video-based synchronization analyzes the trajectories of moving points observed across multi-view videos and estimates the time shift between image sequences without the need for manual intervention. This often minimizes an error function by imposing epipolar constraints on point correspondences established in advance~\cite{albl2017two}. Additionally, the movement of humans and their structural characteristics, such as rhythm and articulation, have been demonstrated flexibility as reference targets for temporal alignment in multi-view synchronization~\cite{Takahashi18,elhayek2012feature}. Recently, deep neural networks have been shown significant potential in aligning unsynchronized video clips captured from different camera views~\cite{Liu_Ai_Xing_Li_Wang_Tao_2024,wu2019multi,jenni2020self}.

\subsubsection{Spatiotemporal calibration}

Simultaneous estimation of spatial and temporal transformations from multi-view videos has been achieved using multimodal data: audio and visual features~\cite{hasler2009markerless}. This method estimates camera poses using Structure-from-Motion (SfM) based on static feature points in the background. It also aligns temporal differences between image sequences by analyzing audio signals. Vision-based spatiotemporal alignment commonly relies on trajectories of dynamic objects to associate multiple views and to minimize geometric projection errors within joint optimization schemes~\cite{caspi2006feature,albl2017two,Takahashi18}. Additionally, motion priors on specific targets can be applied as temporal constraints of the optimization.  In particular, the physics-based motion priors of dynamic human movements and static background facilitate spatiotemporal calibration~\cite{Vo2020SpatiotemporalBA}.

These methods address the challenges of multi-person associations, camera calibration, and temporal alignment separately. In this paper, we show that these three coupled sub-problems can be solved as a single registration problem by utilizing sets of 3D joint trajectories obtained from dynamic human movements and converting them to a unit sphere rather than the appearance-based association.

\section{Two-view registration}
Fig.~\ref{fig:overview} provides an overview of the entire pipeline. Given a set of multiview videos of freely moving people as input, we perform extrinsic camera calibration, synchronization, and association across multiple views. We tackle these challenges through a single registration problem between two views, \ie, source and target, using two sets of 3D joint trajectories obtained from human movements.  We transform the 3D joint trajectories into 3D points on a unit sphere.  Finding the best rotation and a temporal offset between two 3D point sets on a unit sphere yields those of the two views directly. 
Once the relative rotations, temporal offsets, and associating persons between views are estimated, the existing method~\cite{Hartley00} is used to determine a translation. Finally, a non-linear optimization method refines them in a single system by a non-linear optimization method.

\subsection{Problem Formulation}
Suppose two static cameras $c$ and $c^{\prime}$ capture $N_p$ and $N_{p^{\prime}}$ moving people in a scene. Our two-view registration determines a set of parameters  $\langle R, \mathbf{t}, \delta, A \rangle$ representing a 3D rotation $R$ and translation $\mathbf{t}$, a relative temporal offset $\delta$ between two image sequences, and an association matrix $A$. %
The association matrix, a binary matrix $N_p \times N_{p^{\prime}}$, identifies matches between the individuals captured in the two views whose element is 1 where the respective individuals are matched and 0 otherwise.

For each camera view, we use an off-the-shelf monocular 3D human pose estimator to obtain 3D poses~\cite{alphapose,motionbert2022}. A 3D pose of person $p$ can be converted into a set of 3D-oriented points  $o=\braket{\mathbf{x}_{p,\tau}^{(s)}, \mathbf{v}_{p,\tau}^{(s)}}$, where  $\mathbf{x}_{p,\tau}^{(s)}\in \mathbb{R}^3$  denotes the position of the joint $s$ and $\mathbf{v}_{p,\tau}^{(s)} \in \mathbb{R}^3$ indicates the orientation of the joint $s$ relative to its parent joint at frame $\tau$. These orientations can be represented as a set of points on a unit sphere.
The goal of our two-view registration is to find the best $R$, $\mathbf{t}$, $\delta$, and $A$ which minimizes
\begin{equation}
\sum_{p, p' \in A}
\sum_{\tau}
\sum_{s}
\left[\mathbf{v}_{p,\tau}^{(s)} - R\mathbf{v}^{\prime(s)}_{p',\tau+\delta}\right],
\end{equation}
and
\begin{equation}
\sum_{p, p' \in A}
\sum_{\tau}
\sum_{s}
\left|{\mathbf{x}_{p,\tau}^{(s)}}^\top[\mathbf{t}]_{\times}R\mathbf{x}^{\prime(s)}_{p',\tau+\delta}
\right|^2,
\end{equation}
where $p, p' \in A$ denotes each of the corresponding persons $p$ and $p'$ defined by $A$.

\subsection{Simultaneous Calibration and Association}
This section introduces our calibration algorithm for $R$ and $A$ first for simplicity.  The complete algorithm involving $\mathbf{t}$ and $\delta$ is then introduced later.
Suppose we have two sets of 3D orientations from each view over $T$ frames. Here, we assume that each person in each view is tracked over $T$ frames by a conventional method~\cite{alphapose}.
We formulate the registration of these two sets in a probabilistic way that introduces latent variables for association.

For the source view, the 3D orientations of each joint $s$ for $T$ frames are statistically modeled by fitting a von Mises Fisher (vMF) distribution parameterized by its mean direction $\mu_s$ and concentration parameter $\kappa$ as $f_{\mathbf{vMF}}(\mathbf{v};\mu,\kappa)=C(\kappa)\exp(\kappa \mu^{\top} \mathbf{v})$, where $C(\kappa)$ is the normalization constant dependent on $\kappa$.
\textcolor{black}{We fit a vMF distribution for each joint of each person $p'$ in the source view within the $T$ frames, based on the maximum likelihood estimate (MLE) and model the source observation as a mixture of vMFs with mixing proportion $\pi_{p'}$ per person.} 
This model serves as the basis for registering the observed data from the target view. We employ an EM-like alternating optimization strategy which optimizes $R$ and $A$~\cite{Moon1996TheEA}. This allows us to iteratively update the estimates between the spatiotemporal alignment and soft assignments for associating multi-person across views.
To simplify equations, we consider only for the orientation of a single body part hereafter and omit $(s)$ in the equations.

We first define the log-likelihood function over the association $A$ conditioned on the target observation $\mathcal{Y}$ and the parameters $\theta=\{R, \{\pi_{p}\}_{p=1}^{N_{p^{\prime}}}\}$ as $\log p(\mathcal{Y}, A \mid\ \theta)$, where
$\mathcal{Y}$ is given as a set of 3D orientations as $\mathcal{Y} = \{\mathbf{v}_{p,\tau} \mid  p=1, \ldots \, N_p, \tau = 1, \ldots, T \}$.
Our goal is to maximize the log-likelihood of observed data involving the latent variables.  That is, we align the target orientations to a mixture of vMF distributions defined by persons' joints in the source view,
by applying a single rotation $R$.

If the association $A$ were given,
the log-likelihood can be calculated as
\begin{equation}
\log p(\mathcal{Y}, A \mid\ \theta)=\log \prod_{p,p' \in A}\prod_{\tau=1}^{T} \Big\{\pi_{p'}f_{\mathbf{vMF}}(\mathbf{v}_{p,\tau};R\mu_{p'},\kappa_{p'})\Big\}. \label{eq:hard_ll}
\end{equation}
Instead of directly optimizing the binary assignment matrix $A$, we introduce a soft assignment using the posterior probability, commonly referred to as \textit{responsibility}.
The responsibility $\gamma_{p,p'}$ is given by
\begin{equation}
\begin{split}
\gamma_{p,p'} &= \frac{\Pi_{\tau=1}^{T} \pi_{p'}f_\mathrm{vMF} (\mathbf{v}_{p,\tau}; R\mu_{p'}, \kappa_{p'})}
{\sum_{p'=1}^{N_{p'}} \Pi_{\tau=1}^{T} \pi_{p'}f_\mathrm{vMF} (\mathbf{v}_{p,\tau}; R\mu_{p'}, \kappa_{p'}) } .
\label{eq:gamma}
\end{split}
\end{equation}
By rewriting the log-likelihood in Eq.~\eqref{eq:hard_ll} with this soft assignment, we have
\begin{equation}
\sum_{p'=1}^{N_{p'}}
\sum_{p=1}^{N_{p}}
\sum_{\tau=1}^{T} \gamma_{p,p'} \left(\log \pi_{p'}+ \log f_{\mathbf{vMF}}(\mathbf{v}_{p,\tau};R\mu_{p'},\kappa_{p'})\right). \label{eq:soft_ll}
\end{equation}

The partial derivative of Eq.~\eqref{eq:soft_ll} \wrt $R$ is given as
\begin{equation}
\begin{split}
   \nabla_{R} &= \frac{\partial }{\partial R } \sum_{p'=1}^{N_{p'}}
   \sum_{p=1}^{N_{p}}
   \sum_{\tau=1}^{T}
   \gamma_{p,p'}(\kappa_{p'} (R\mu_{p'})^{\top} \mathbf{v}_{p,\tau}) \\
    &=
   \sum_{p'=1}^{N_{p'}}
   \sum_{p=1}^{N_{p}}
   \sum_{\tau=1}^{T}
   \gamma_{p,p'}(\kappa_{p'} ([-R\mu_{p'}]_{\times})^{\top} \mathbf{v}_{p,\tau}).
   \label{eq:pd_r}
\end{split}
\end{equation}
We use the Lie algebra parameterization $\mathbf{\omega} \in \mathbb{R}^3$ along with the perturbation model of infinitesimal rotation $\frac{\partial (R\mathbf{\mu})}{\partial \omega } \approx \frac{\partial (R\mathbf{\mu})}{\partial \Delta \omega } = -[R\mathbf{\mu}]_{\times}$, which makes the optimization with respect to rotation more efficient and robust, particularly in avoiding common issues associated with other rotation parameterizations~\cite{solà2021micro}. 

The partial derivative of Eq.~\eqref{eq:soft_ll} \wrt $\pi_{p'}$ is $\nabla_{\pi_{p'}}= \frac{\partial }{\partial \pi_{p'} } 
   \sum_{p'=1}^{N_{p'}}
   \sum_{p=1}^{N_{p}}
   \gamma_{p,p'} \log \pi_{p'}.$ The closed-form solution of the optimal $\pi_{p'}$ can be obtained using the normalizing constraint $\sum_{p'=1}^{N_{p^{\prime}}}\pi_{p'}= 1$ and a Lagrange multiplier $\lambda$ as 
\begin{equation} 
\begin{split}
   \frac{\partial }{\partial \pi_{p'} } \left( 
   \sum_{p'=1}^{N_{p'}}
   \sum_{p=1}^{N_{p}}
   \gamma_{p,p'} \log \pi_{p'}+\lambda
\left(\sum_{p'=1}^{N_{p^{\prime}}}\pi_{p'}-1\right)\right)=0.
   \label{eq:pd_pi}
\end{split}
\end{equation}
The closed-form solution of Eq.~\eqref{eq:pd_pi} for $\pi_{p'}$ is:
\begin{equation} 
\begin{split}
\pi_{p'} = \frac{\sum_{p=1}^{N_{p}} \gamma_{p,p'}}{\sum_{p'=1}^{N_{p^{\prime}}}\sum_{p=1}^{N_{p}} \gamma_{p,p'}},
 \label{eq:closed_form_pi}
\end{split}
\end{equation}
which allows us to optimize the responsibility $\gamma_{p,p'}$ under the current estimate of $\hat{R}$.

Using Eqs.~\eqref{eq:gamma},~\eqref{eq:pd_r}, and ~\eqref{eq:closed_form_pi}, our optimization process alternates between finding the parameters $\theta$ and calculating the responsibility $\gamma_{p,p'}$ until convergence. To obtain the association $A$,
we solve the bipartite  matching~\cite{Bijsterbosch2010SolvingTR} of the graph defined by the target $N_p$ and the source $N_{p'}$ persons with its edge weight
$C_{p,p'} = %
\sum_{\tau=1}^{T} \arccos{({\mathbf{v}_{p,\tau}},\hat{R}\mathbf{v}_{p',\tau}^{\prime})}. \label{eq:bp_cost}$

Once $R$ and $A$ are obtained, we can estimate $\mathbf{t}$ by solving a system of linear equations derived from the epipolar constraint~\cite{Hartley00}.

\subsection{Simultaneous Calibration, Association, and Synchronization}

The simultaneous calibration and association algorithm in the previous section can be extended to estimate the temporal offset $\delta$ in a unified manner.

By populating the vMFs at different offsets on the source side, our algorithm automatically finds the best rotation $R$, together with the responsibility $\gamma_{p,p'}$ which represents both the association and the offset in an entangled manner.

To estimate the association and the offset, instead of disentangling $\gamma_{p,p'}$ straightforwardly, in practice, we employ a brute-force search which examines all possible offsets between each camera pair.  This is because the possible temporal offset is bounded in 1D discrete space, and examining the plausibility of each offset is rather simple.

\subsection{Outlier Removal}\label{sec:RANSAC}

While our algorithm can handle occlusions and non-overlapping observations automatically by setting the responsibility zeros in theory, 3D poses lookalike each other can result in wrong matches.

We introduce an additional outlier removal process using RANSAC.  We randomly select $\hat{N}_{p'} < N_{p'}$ source-side persons as those who are hypothesized to be visible also in the target side, and estimate the $R, \mathbf{t}, \delta$ and $A$ based under that hypothesis.  By using the metric in Eq.~\eqref{eq:bp_cost}, we can find the best $R, \mathbf{t}, \delta$ and $A$ from all hypotheses.

In this process, $\hat{N}_{p'}$ balances a trade-off between the combinatorial complexity and matching ambiguity.  Larger $\hat{N}_{p'}$ makes the calibration process less ambiguous while it increases the RANSAC space and also increases the risk of having the hypothesized persons invisible in the target side.  In the evaluations, we empirically set $N_{p'} = 2$ to balance these two aspects.

\section{Multi-view Integration and Optimization}\label{sec:BA}

We unify the results of pairwise calibrations and multi-person associations from a multi-view setup with $N_C$ cameras into a single optimization framework. Pose graph optimization (PGO)~\cite{kummerle2011g} using an essential graph over the set of all possible paired transformations $\mathcal{T}=\{R_{i,j}, \mathbf{t}_{i,j} \mid 1 \leq i,j \leq N_C, i \neq j\}$ unifies the transformations to ensure global consistency across all camera views. We use a spanning tree graph to integrate the multi-person associations $\mathcal{A}=\{A_{i,j}\mid 1 \leq i,j \leq N_C, i \neq j\}$, which ensures a consistent set of global correspondences and cycle consistency for subsequent optimization.

In order to optimize the camera calibration, synchronization, and association, we introduce a spatiotemporal bundle adjustment (STBA).  The key in our STBA is to render the temporal offset optimization as a reprojection error minimization. By assuming the human motion to be linear in a short period of time, we can synthesize the 3D position of each joint under a hypothesized temporal offset as $\mathbf{X}^{c}_{s}=\mathbf{X}_s+ \delta^{c} \mathbf{V}_{s}$, where $\mathbf{X}_s$ and $\mathbf{V}_s$ represent the triangulated 3D point and the 3D motion for $s$-th joint, respectively, and $\delta^{c}$ denotes the temporal offset for camera $c$, respectively. We can evaluate this by measuring reprojection errors. This linear approximation allows us to integrate the $R$, $\mathbf{t}$, and $\delta$ refinement into a single non-linear optimization problem under a current association $A$.

Once $R$, $\mathbf{t}$, and $\delta$ are optimized, we can verify the association by verifying the reprojection error for each person.  We discard associations with large errors, and iterate this STBA and outlier removal until convergence for robust estimation of $R$, $\mathbf{t}$, $\delta$, and $A$.

\section{Experiments}

In this section, we present both quantitative and qualitative evaluations of the proposed method, focusing on its applications in various scene configurations using synthetic and real data. The intrinsic calibration was performed in advance. We benchmarked against the state-of-the-art approaches to the point-set registration problem, ARCS \cite{Peng2022ARCSAR} and extrinsic camera calibration in a multi-view system involving human subjects, referred to as ReID-Calib \cite{yan2021wide-baseline}. Since the original implementation of ReID-Calib is not publicly available, we reimplemented it using a state-of-the-art person reidentification model \cite{chen2023beyond}. By incorporating temporal alignment into our method, we compare it with the existing approach~\cite{Vo2020SpatiotemporalBA}. The experiments were approved by the Research Ethics Committee of the Graduate School of Informatics, Kyoto University (KUIS-EAR-2020-002).

\subsection{Experimental Setups}

For our evaluation metrics, we utilize Riemannian distance~\cite{moakher2002means} to quantify rotation errors as $E_{R}$, root mean square error (RMSE) to measure translation errors as $E_{\mathbf{t}}$ up to scale, and reprojection errors $E_{\mathrm{2D}}$ in pixels. Here $E_{\mathbf{t}}$ is normalized by the ground truth distance between the first two cameras. Additionally, mean absolute error (MAE) is used to measure the errors in temporal offset as $E_{\mathrm{\delta}}$ in frame scale, indicating the time shift between two sequences. To evaluate the performance of establishing correspondences, we employ the precision metric $\mathcal{P}=\frac{\mathrm{Correct\: matches}}{\mathrm{Estimated \:matches}}$. For real datasets without ground-truth person associations, we assess \(\mathcal{P}\) through geometric consistency using multiple views.  We calculate reprojection errors after triangulating points based on the matches established across views using the ground truth of camera poses. Correct matches are defined by the criterion that requires the reprojection error to be within five pixels. 

For our experiments, we uniformly subsample the sequences within each scenario and apply random initial rotations. Trackings are initiated from each of subsampled frames for $T$ frames.
We set $T$ between 5 and 15 frames, depending on the scenes. In the following experiments with unsynchronized setups, the temporal offset is set to $+5$ frames relative to the first camera with a search range of temporal candidates of $\pm 10$ frames. The optimization begins with initial offsets set to zero.

\begin{table*}[ht]

\begin{center}
\centering
\resizebox{\textwidth}{!}{
\begin{tabular}{ll|cccc|cccc|cccc}
\toprule
 {}& {} & \multicolumn{4}{c|}{CMU Panoptic (\textit{160906\_pizza1})~\cite{Joo_2017_TPAMI}} & \multicolumn{4}{c|}{ZJU-MoCap (\textit{soccer1\_6})~\cite{shuai2022novel}} & \multicolumn{4}{c}{MMPTRACK (\textit{industry\_safety\_2})~\cite{han2021mmptrack}} \\
 {} & {} &                            
 $E_R $[\si{\radian}] & $E_\mathbf{t}$ & $E_\mathrm{2D}$[\si{px}]& $\mathcal{P} \uparrow$ & 
 $E_R $[\si{\radian}] & $E_\mathbf{t}$ & $E_\mathrm{2D}$[\si{px}]& $\mathcal{P} \uparrow$ & 
 $E_R $[\si{\radian}] & $E_\mathbf{t}$ & $E_\mathrm{2D}$[\si{px}]& $\mathcal{P} \uparrow$ \\
 
\midrule

\multirow{4}{*}{ARCS~\cite{Peng2022ARCSAR}}  
&  PR
& 1.625 & 1.737 & 625.409  & 0.036 
& 2.303 & 1.839 & 387.707  & 0.045 
& 2.341 & 1.89 & 175.14  & 0.072 
\\  &   MI
& 0.605 & 1.078 & 1113.369  & 0.069 
& 1.573 & 0.556 & 19741.342  & 0.068 
& 1.334 & 0.83 & 289.307 & 0.098  
\\&  BA
& 0.564 & 1.075 & 369.306  & 0.089 
& 1.599 & 0.506 & 101.335  & 0.074 
& 1.346 & 0.829 & 76.991  & 0.107
\\  & IBA          
& 0.573 & 1.083 & 34.383  & 0.098 
& 1.596 & 0.497 & 12.481  & 0.068 
& 1.334 & 0.811 & 13.817  & 0.101 \\
 
\midrule \multirow{4}{*}{ReID-Calib~\cite{yan2021wide-baseline}}  
&  PR
& 2.078 & 1.295 & 1170.047 & 0.035 
& 2.142 & 1.428 & 815.481  & 0.133 
& 2.075 & 1.2 & 925.881 & 0.173 
\\& MI
& 1.726 & 1.066 & 741.103  & 0.102 
& 1.355 & 0.415 & 662.632  & 0.257 
& 0.77 & 0.727 & 287.999  & 0.288 \\
&  BA       
& 1.56 & 1.209 & 110.689  & 0.11 
& 1.499 & 0.526 & 61.222  & 0.276 
& 1.063 & 0.699 & 35.779  & 0.328\\  
& IBA          
& 1.448 & 1.059 & 59.457  & 0.166 
& 1.465 & 0.487 & 14.711  & 0.357 
& 1.051 & 0.622 & 2.837  & 0.387 \\ 
\midrule \multirow{4}{*}{Ours}  
&  PR                
& 1.603 & 0.883 & 28.568  & 0.552 
& 1.126 & 0.488 & 19.687  & 0.392 
& 1.704 & 0.767 & 18.651  & 0.327 \\
&  MI                  
& 0.498 & 1.0 & 127.16 & 0.59 
& 0.347 & 0.184 & 93.824  & 0.568 
& 0.304 & 0.537 & 53.21 & 0.635 \\
&  BA
& \textbf{0.024} & 0.029 & 4.492  & 0.667 
& 0.072 & 0.075 & 12.665  & 0.636 
& 0.146 & 0.138 & 8.327  & 0.736 \\
 
& IBA
& 0.027 &\textbf{ 0.023} & \textbf{3.875 }&  \textbf{0.706} 
& \textbf{0.006} & \textbf{0.004} & \textbf{3.778} &  \textbf{0.722 }
& \textbf{0.024} &\textbf{ 0.033} & \textbf{2.415} &  \textbf{0.896}  \\ 

 \bottomrule
\end{tabular} 
}
\end{center}
\vspace{-3mm}
\caption{Evaluations of extrinsic calibration and person association in pre-synchronized video scenarios.}
\vspace{-3mm}
\label{table:ECalib}
\end{table*}

\begin{table*}[]
\centering
\resizebox{\textwidth}{!}{
\begin{tabular}{cl|ccc|ccc|ccc}
\toprule
{} &{}& \multicolumn{3}{c|}{CMU Panoptic~\cite{Joo_2017_TPAMI}} & \multicolumn{3}{c|}{ZJU-MoCap~\cite{peng2021neural,shuai2022novel}} & \multicolumn{3}{c}{MMPTRACK~\cite{han2021mmptrack}} \\
{} &{}& 
\textit{160906\_pizza1} &\textit{170915\_toddler5} &\textit{161029\_hands1} &
\textit{soccer1\_6} &\textit{ballet} &\textit{boxing} &
\textit{industry\_safety\_2} &\textit{cafe\_shop\_3} &\textit{lobby\_3} \\ \midrule

\multicolumn{2}{c|}{\# of People} &
6&3&3&
6&2&2&
6&7&7 \\ 

\multicolumn{2}{c|}{Camera ID} &
[2,12,21,28]&[12,15,22,28]&[1,4,12,18]&
[1,3,5,7]&[4,7,16,18]&[5,7,8,9]&
[1,2,3,4]&[1,2,3,4]&[1,2,3,4] \\ \midrule

\multirow{4}{*}{Vo\etal~\cite{Vo2020SpatiotemporalBA} }
& $E_R$[\si{\radian]}

&0.03&-&0.006
&-&-&0.013
&-&-&- \\
& $E_\mathbf{t}$  
&0.062&-&0.014
&-&-&0.014
&-&-&- \\
& $E_\mathrm{2D}$[\si{px}]  
&6.238&-&6.503
&-&-&3.814
&-&-&- \\
& $E_{\delta}$[\si{frame}]
&\textbf{0}&-&\textbf{0}
&-&-&\textbf{0}
&-&-&- \\ \midrule
\multirow{5}{*}{STBA}
& $E_R$[\si{\radian]} 
& 0.311& 0.227& 0.005
& 0.071& 0.015& \textbf{0.007}
& 0.086& 0.26& 0.71\\
 
 & $E_\mathbf{t}$ 
& 2.815& 0.121& 0.012
& 0.039& 0.027& \textbf{0.009}
& 0.081& 0.355& 0.515\\

 & $E_\mathrm{2D}$[\si{px}]  
& 13.983& 7.856& 6.155
& 12.531& 7.994& 4.111
& 5.157& 9.71& 24.153\\

 & $E_{\delta}$[\si{frame}] 
& 4& 3& \textbf{0}
& 6& 1& \textbf{0}
& 6& 4& 8\\

 & $\mathcal{P} \uparrow$  
& 0.628& 0.572& 0.376
& 0.621& 0.403& 0.67
& 0.852& 0.628& 0.607\\
 \midrule
\multirow{5}{*}{ISTBA}
& $E_R$[\si{\radian]} 
& \textbf{0.012}& \textbf{0.012}& \textbf{0.004}
& \textbf{0.007}& \textbf{0.013}& 0.014
& \textbf{0.016}& \textbf{0.025}& \textbf{0.078}\\
 & $E_\mathbf{t}$ 
& \textbf{0.036}& \textbf{0.018}& \textbf{0.009}
& \textbf{0.006}& \textbf{0.025}& 0.014
& \textbf{0.021}& \textbf{0.109}& \textbf{0.047}\\

 & $E_\mathrm{2D}$[\si{px}]  
& \textbf{2.984}& \textbf{4.28}& \textbf{5.481}
& \textbf{3.282}& \textbf{5.732}& \textbf{3.434}
& \textbf{1.461}& \textbf{1.307}& \textbf{1.129}\\

 & $E_{\delta}$[\si{frame}] 
& 1& \textbf{1}& \textbf{0}
& \textbf{1}& \textbf{0}& \textbf{0}
& \textbf{0}& \textbf{1}& \textbf{1}\\

 & $\mathcal{P} \uparrow$  
& \textbf{0.75}& \textbf{0.593}& \textbf{0.377}
& \textbf{0.662}& \textbf{0.405}& \textbf{0.671}
& \textbf{0.933}& \textbf{0.875}& \textbf{0.961}\\

\bottomrule

\end{tabular}
}
\vspace{-1mm}
\caption{Evaluations of spatiotemporal calibration and person association in unsynchronized video scenarios. `-' indicates that the evaluation was not applicable due to failure in the initialization step.}
\label{table:STCalib}
\vspace{-6mm}
\end{table*}

\begin{figure}[t]
\centering
\begin{tabular}{@{}c@{}c@{}c@{}c@{}}
& \textit{160906\_pizza1} & \textit{soccer1\_6} & \textit{industry\_safety\_2} \\
\raisebox{4ex}{\rotatebox[origin=c]{90}{Input}} &
\includegraphics[width=0.3\linewidth]{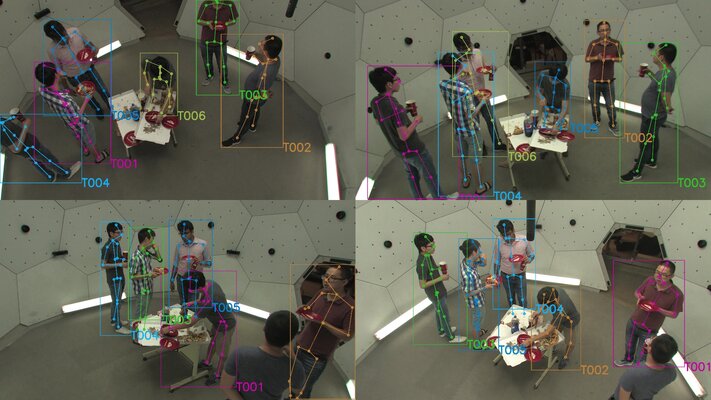} \hspace{-0.3em} &
\includegraphics[width=0.3\linewidth]{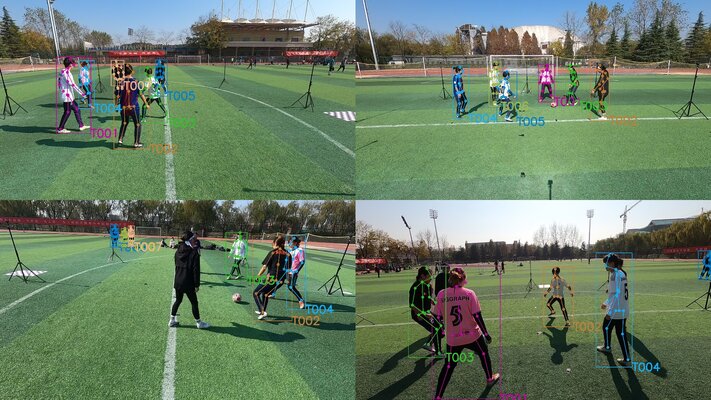} \hspace{-0.3em} &
\includegraphics[width=0.3\linewidth]{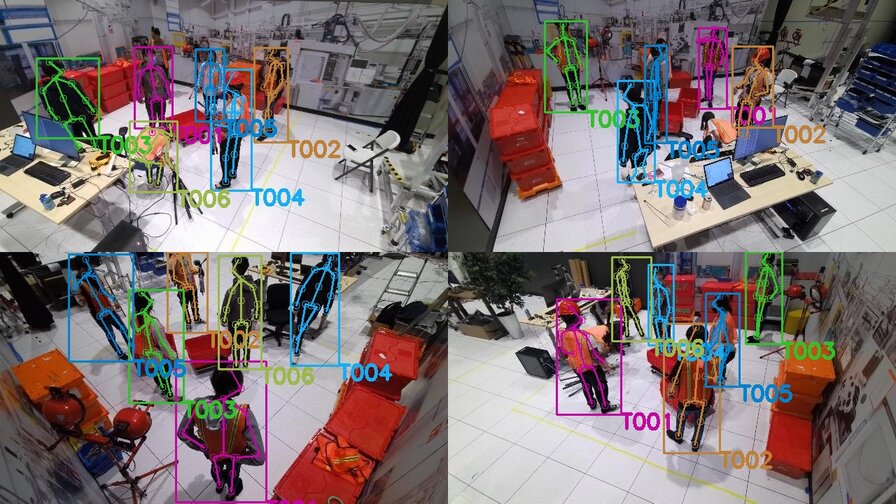} \\
 &
\includegraphics[width=0.3\linewidth]{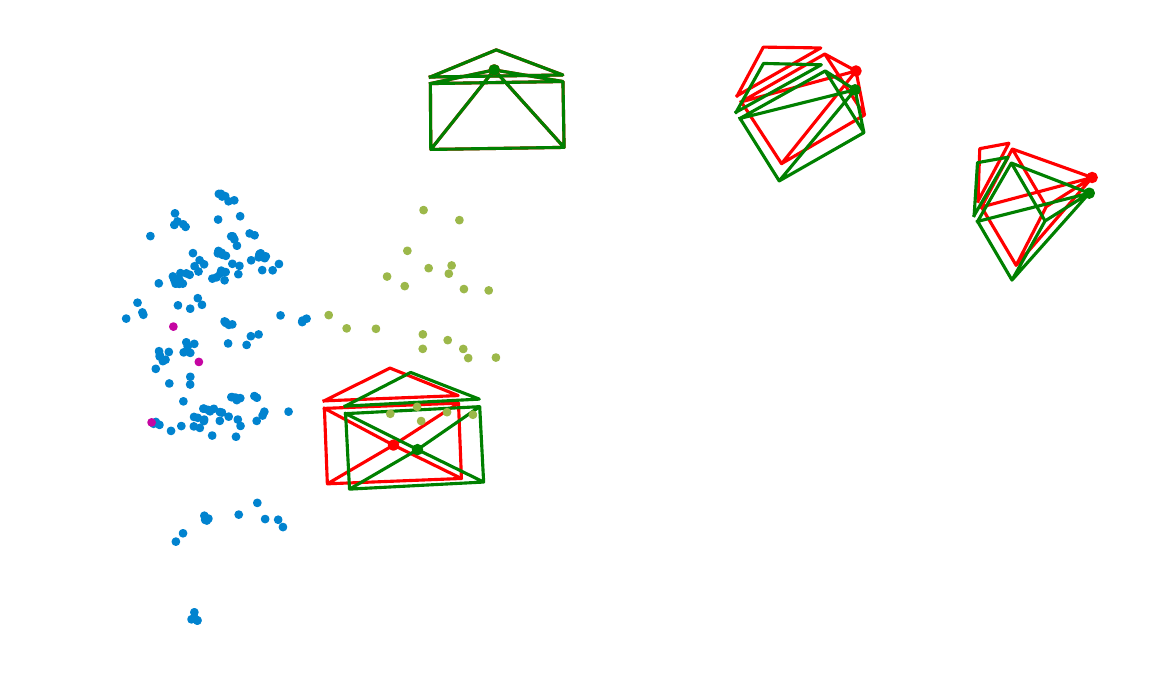} \hspace{-0.2em} &
\includegraphics[width=0.3\linewidth]{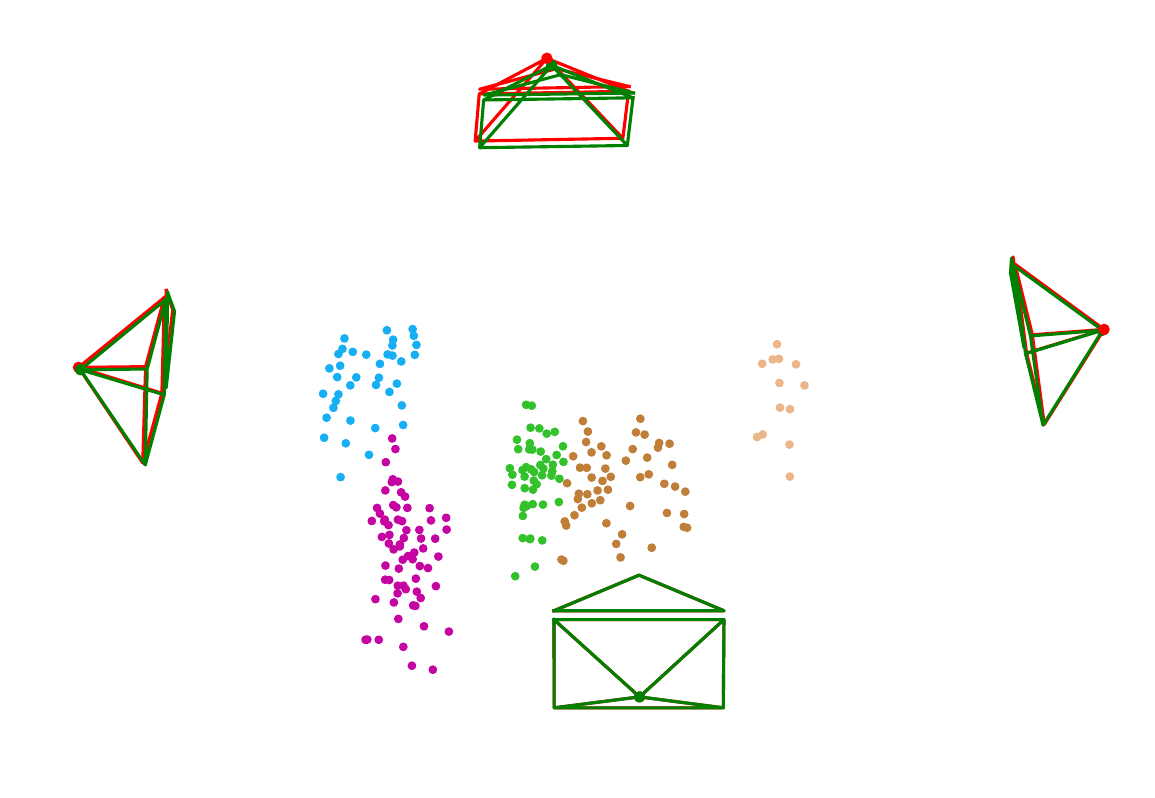} \hspace{-0.2em} &
\includegraphics[width=0.3\linewidth]{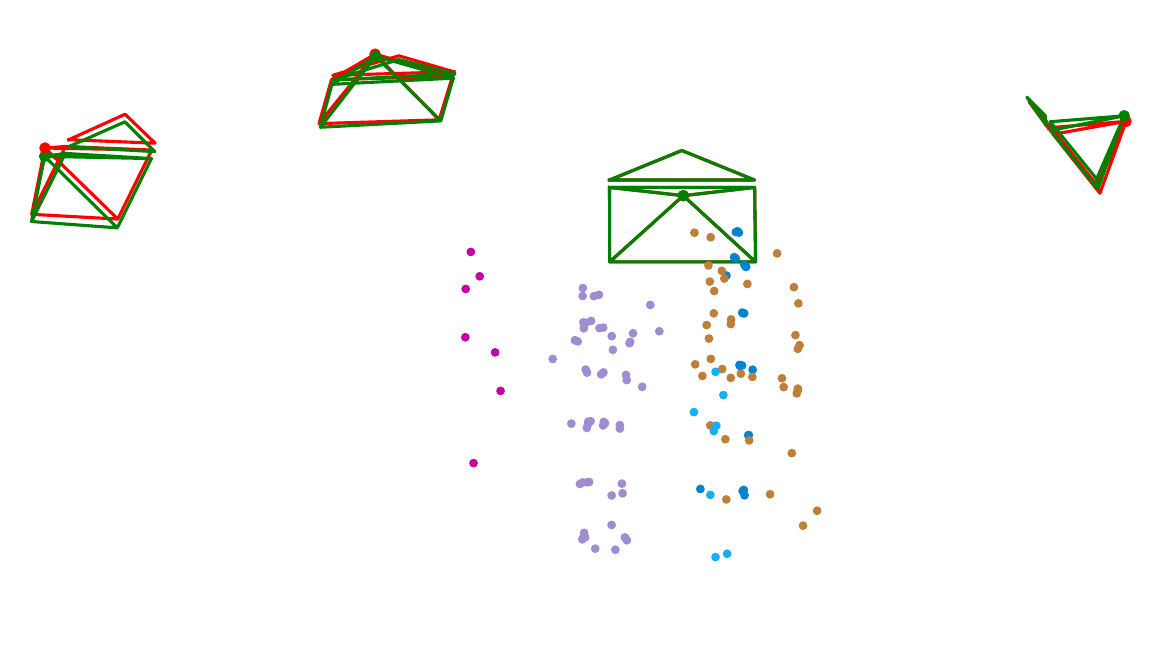} \\
\end{tabular}
\vspace{-3mm}
\caption{Multi-view reconstruction results. Estimated and ground truth camera poses are shown in red and green, respectively, while differently colored 3D points denote individual identification.}
\label{fig:qual_ba}
\vspace{-2mm}
\end{figure}

\begin{figure}[t!]
\centering
\begin{tabular}{@{}c@{}c@{}c@{}}
  \includegraphics[width=.325\linewidth]{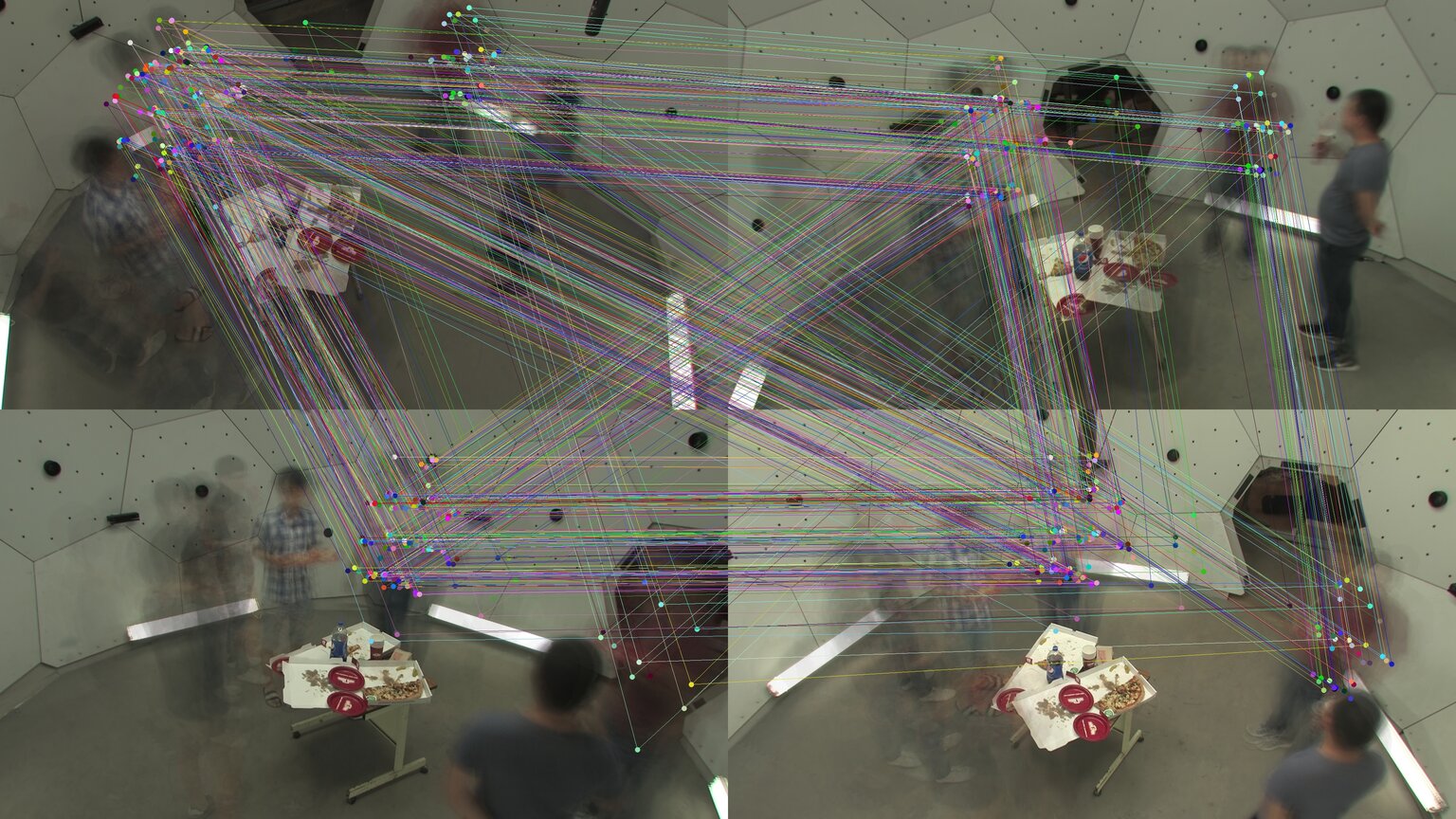} \hspace{-0.2em} &
  \includegraphics[width=.325\linewidth]{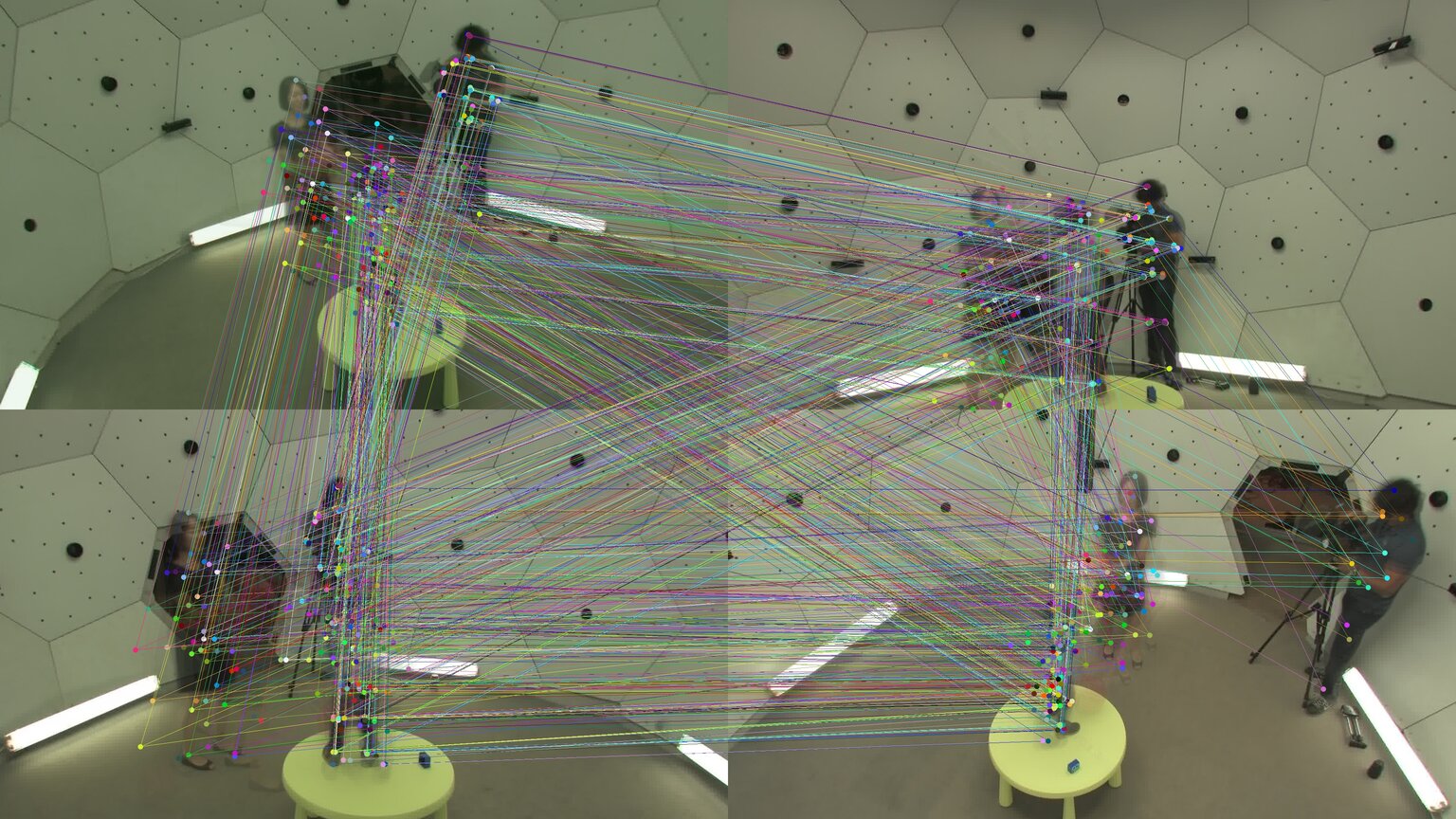} \hspace{-0.2em} &
  \includegraphics[width=.325\linewidth]{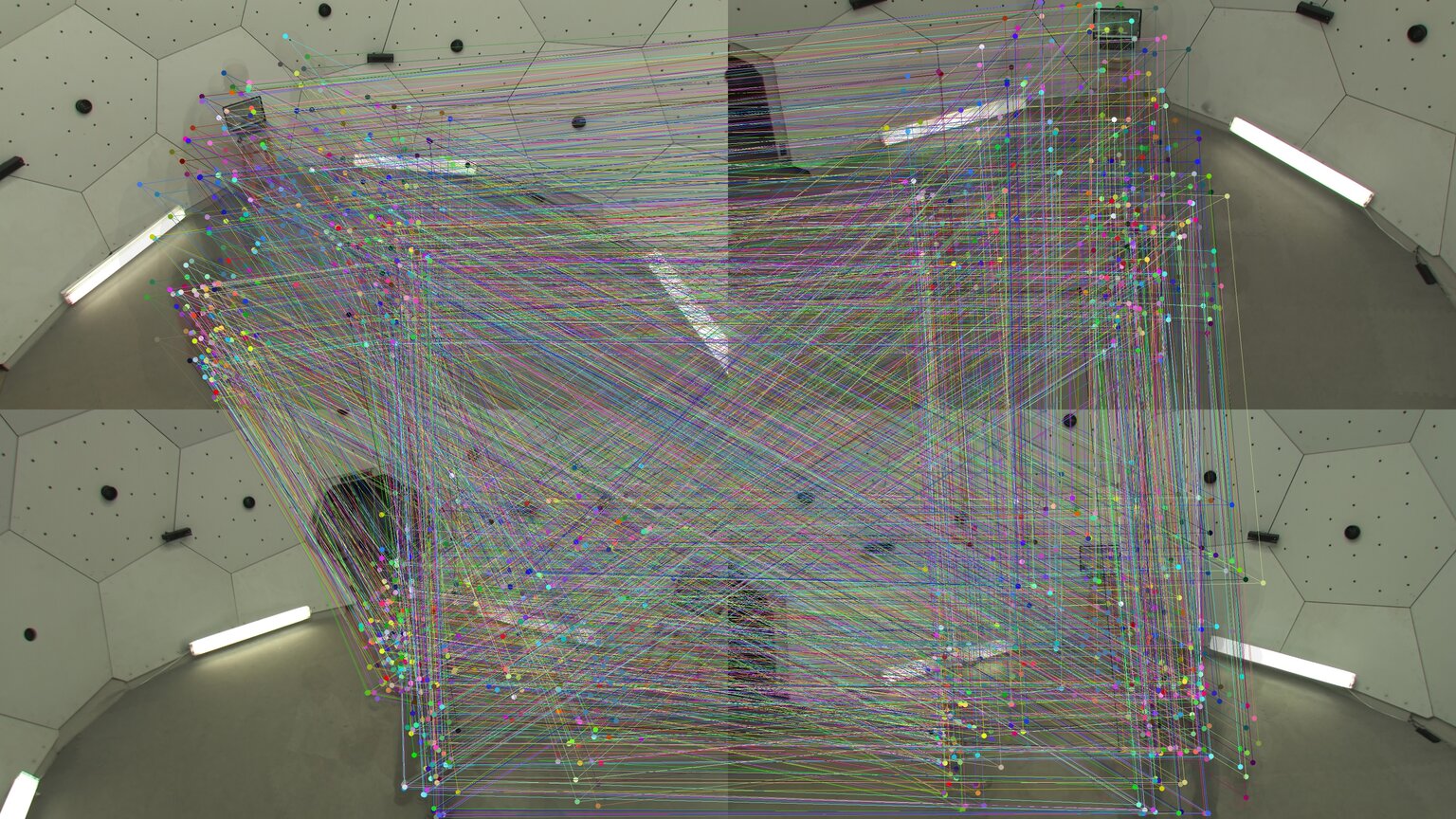} \\
  \vspace{-1.4em} \\
    \includegraphics[width=.325\linewidth]{figures/soccer_matches_new_resized.jpg} \hspace{-0.2em} &
  \includegraphics[width=.185\linewidth]{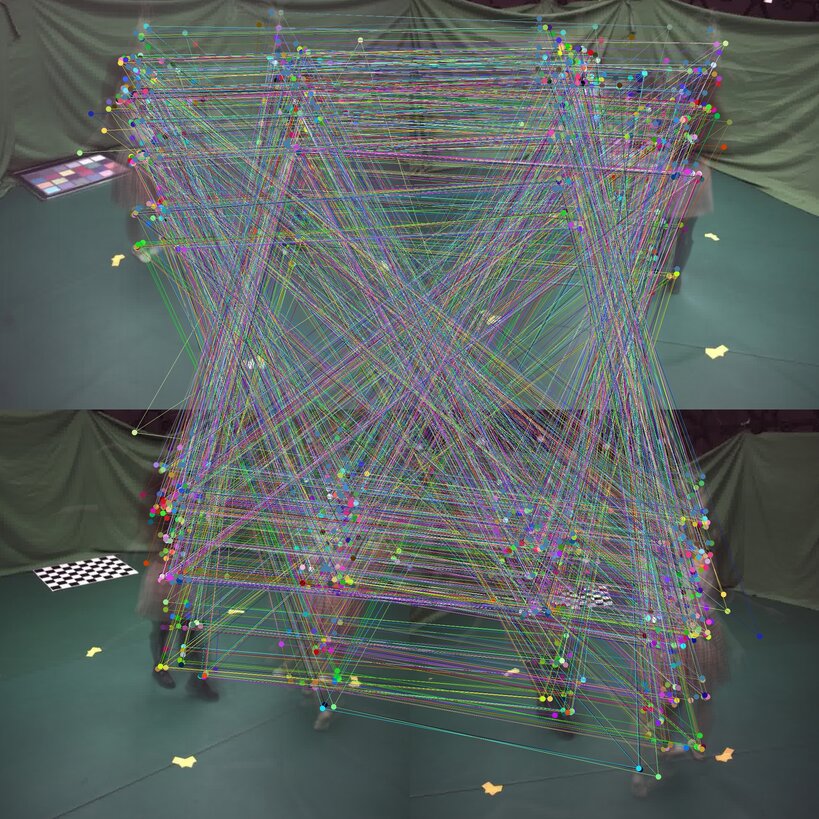} \hspace{-0.2em} &
  \includegraphics[width=.185\linewidth]{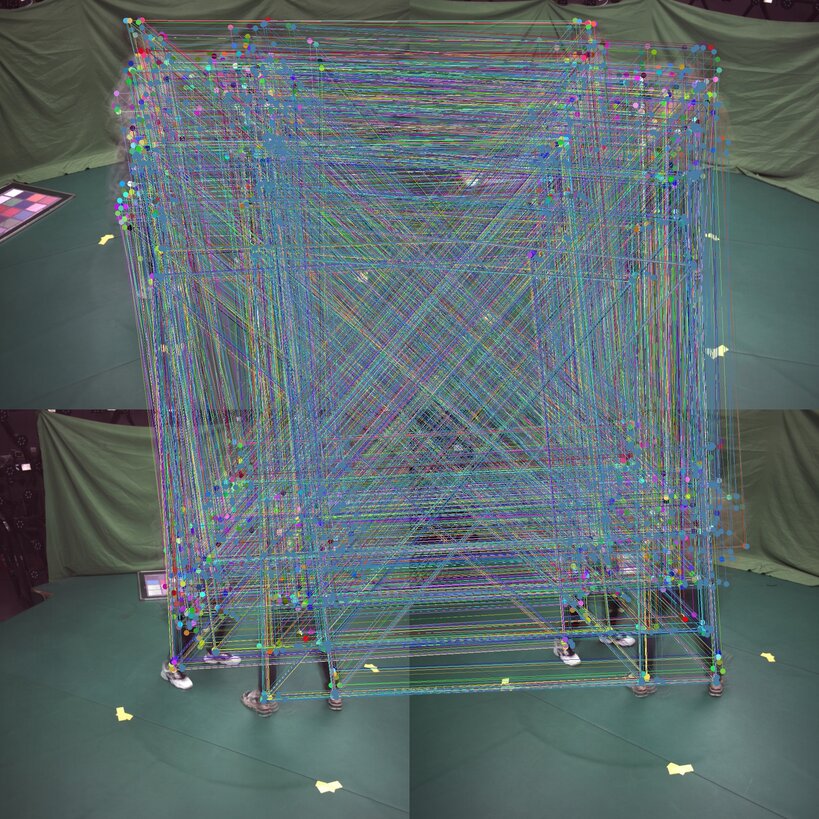} \\
  \vspace{-1.4em} \\
  \includegraphics[width=.325\linewidth]{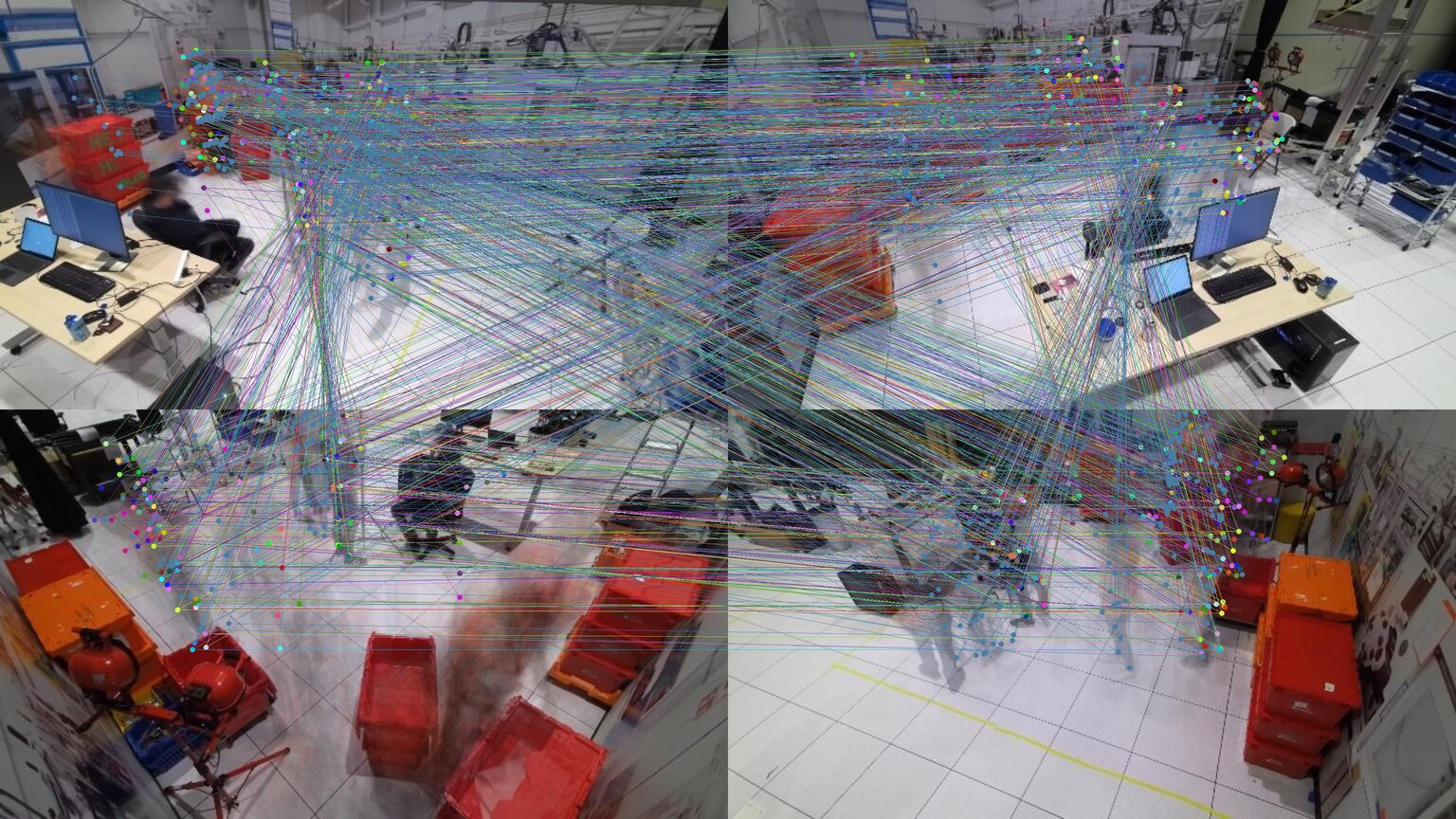} \hspace{-0.2em} &
  \includegraphics[width=.325\linewidth]{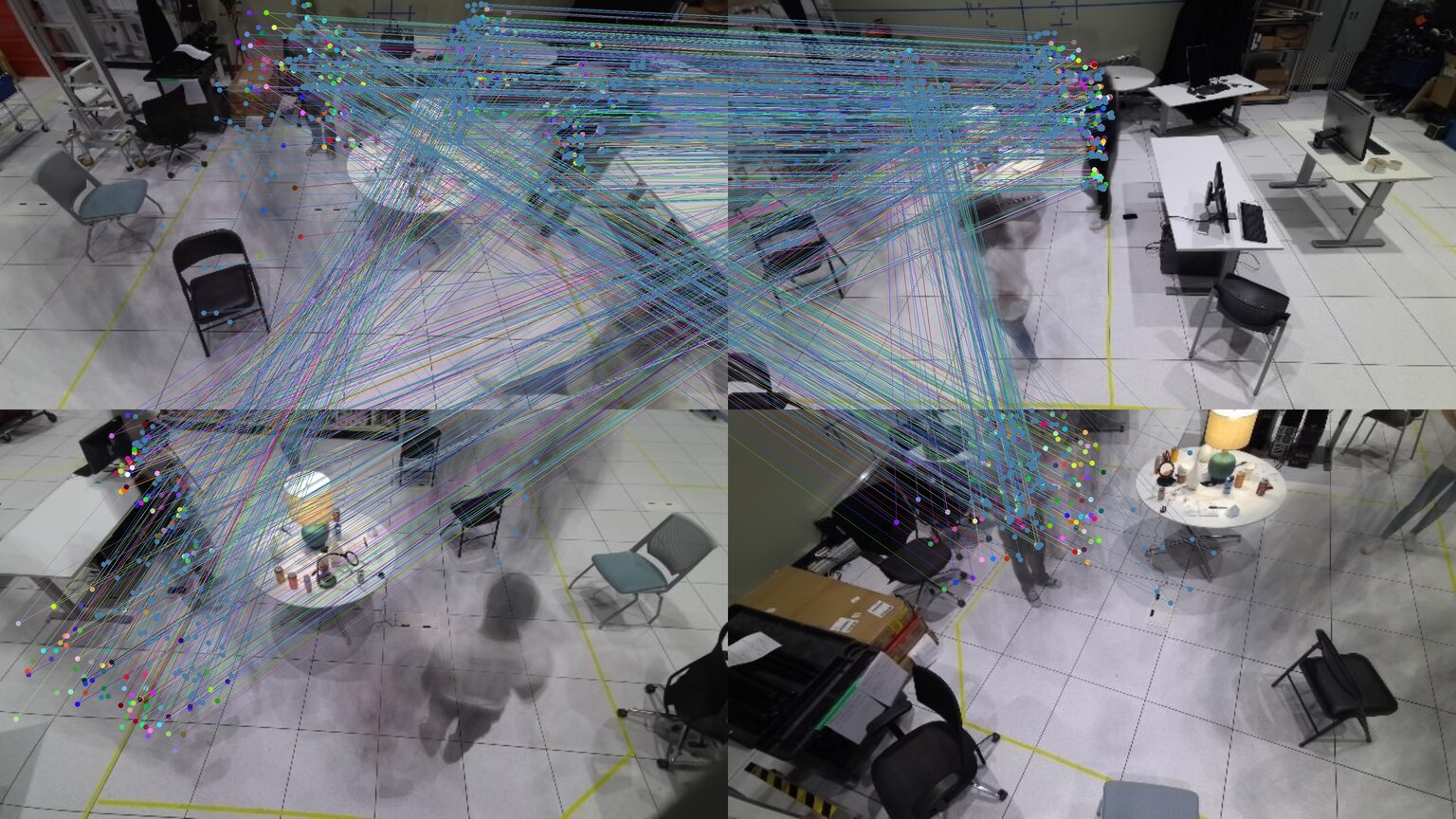} \hspace{-0.2em} &
  \includegraphics[width=.325\linewidth]{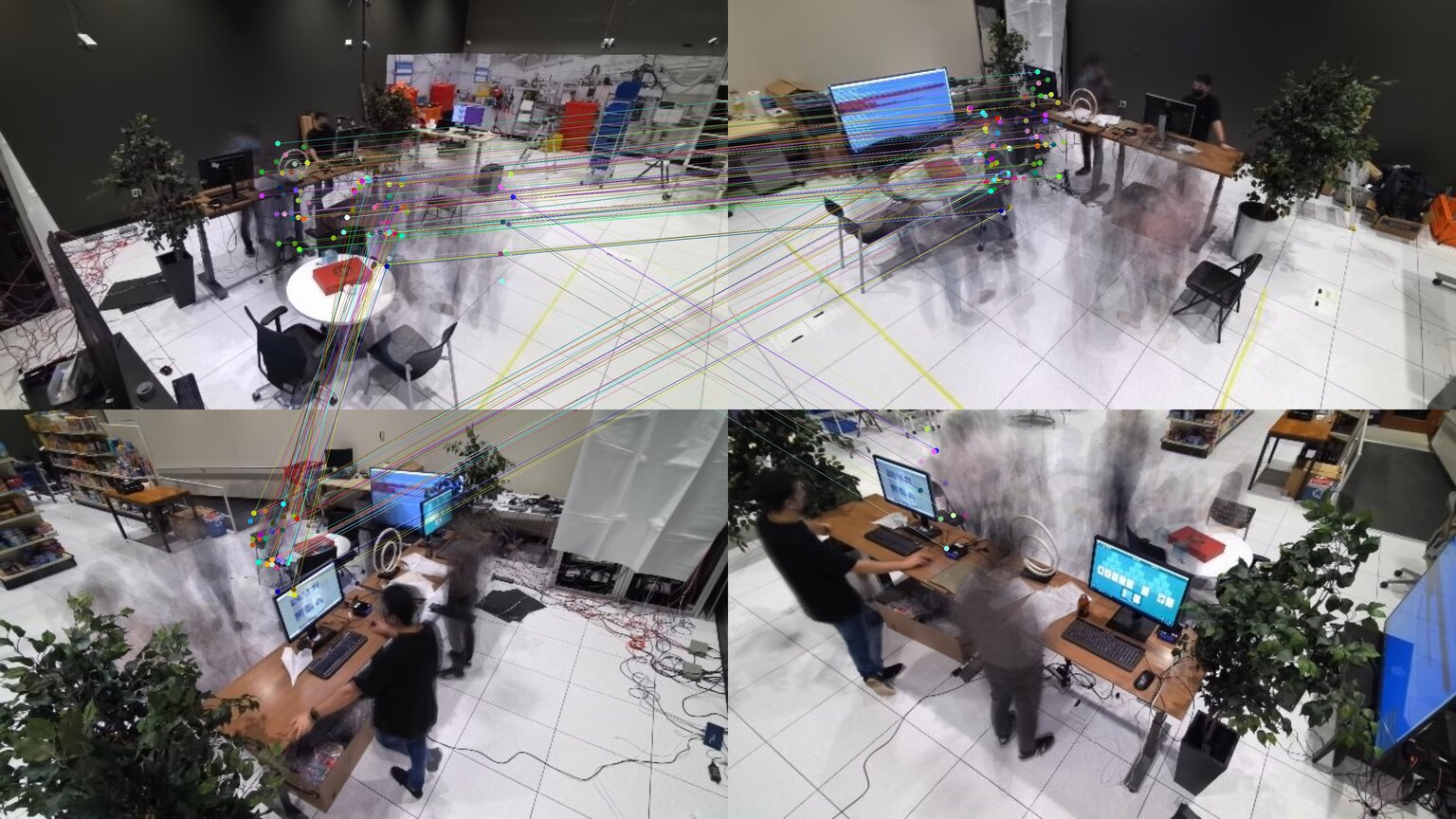} \
\end{tabular}
\vspace{-2mm}
\caption{Cross-view match visualization. The colored lines indicate correspondences between different camera views. }
\label{fig:qual_sba} 
\vspace{-5mm}
\end{figure}

\subsection{Generalizability to Different Scenes}

We validated our algorithm on three datasets:  CMU Panoptic~\cite{Joo_2017_TPAMI}, ZJU-MoCap~\cite{peng2021neural,shuai2022novel}, and MMPTRACK dataset~\cite{han2021mmptrack}, which includes a variety of camera configurations and human activities across both indoor and outdoor environments, as shown in~Fig.~\ref{fig:qual_ba} and Fig ~\ref{fig:qual_sba}. Table~\ref{table:STCalib} provides a summary of these datasets. In each scenario, we selected four cameras, each observing two or more people interacting freely throughout a clip of 300 to 1000 frames. For the CMU Panoptic dataset, we used a set of HD cameras. 

We fully leverage off-the-shelf monocular models to estimate 3D human pose for each person. We used AlphaPose~\cite{alphapose} for 2D detection and tracking of persons with their 2D human poses, and MotionBert~\cite{motionbert2022} for estimating the 3D poses from them. %
Fig.~\ref{fig:qual_ba} shows the results of multi-person pose estimation at the initial frame.  Note that the colors identifying tracked 2D joints are unique to each view and are not shared across views.

Table~\ref{table:ECalib} and Table~\ref{table:STCalib} present the quantitative evaluations.
\subsubsection{Synchronized scenario} In the proposed method, pairwise registration (PR) suffers from larger errors in camera poses due to the inherent uncertainty of 3D human pose estimation in a few pairs of estimation. Using the multi-view integration (MI) with PGO enhances the pose estimation accuracy but also leads to higher reprojection errors due to incorrect matches across multiple views. With the bundle adjustment (BA), however, this issue is resolved by eliminating geometrically inconsistent outliers as described in Sec.~\ref{sec:BA}. We repeat this after BA by setting a threshold three times the median reprojection errors. This Iterative BA (IBA) improves the robustness of establishing the multi-person association and calibrating extrinsic parameters simultaneously. As shown in Fig.~\ref{fig:qual_ba}, the scene reconstructions demonstrate that IBA successfully establishes correspondences and estimates the extrinsic parameters simultaneously in various scenarios despite using off-the-shelf pose estimators. Moreover, the quantitative evaluations show the proposed method outperforms the state-of-the-art baseline approaches in terms of accuracy. This is attributed to treating association and calibration tasks as a single registration problem using 3D human poses rather than as independent problems.    

\subsubsection{Unsynchronized scenario} Due to noisy inputs inherent in 3D human pose estimation, performing pairwise registration with temporal alignment together posed significant challenges. In real-world scenarios, temporal alignment is incorporated into the BA step. As shown in Table~\ref{table:STCalib}, our STBA successfully recovers extrinsic parameters, temporal offsets, and correspondences simultaneously. However, some scenarios still encounter several mismatches in established correspondences due to occlusions or similar poses between individuals. Our iterative STBA (ISTBA) can further refine the parameters by discarding these inconsistent associations found at each round of STBA.  %
As shown in Fig.~\ref{fig:qual_sba}, cross-view matching results demonstrate that the proposed method successfully calibrates and synchronizes multiple cameras while establishing geometrically consistent matches even if the input sequences are not temporally aligned.

For comparison with \cite{Vo2020SpatiotemporalBA}, we reimplemented the method as the original code is publicly unavailable. We used the incremental SfM pipeline~\cite{schonberger2016structure}, to initialize camera poses based on the image features~\cite{Lowe:SIFT} extracted from the initial frames, regardless of background and foreground. The three scenarios succeeded in initialization, while the others failed to find an initial image pair or register other views due to large baselines between views. We applied our STBA to the successful scenario, as it leverages motion priors from moving people in the foreground; however, the person associations were given here—unlike in our approach. After successful initial registration, both methods accurately estimated extrinsic parameters and temporal offsets in multi-camera setups. Our results demonstrate that leveraging moving people in the foreground enhances the reliability of the initialization prior to applying BA. Additionally, the proposed method integrates the estimation of correspondences, streamlining the process and removing reliance on pre-established correspondences.

\begin{figure}[t]
  
\centering
\begin{tabular}{c}

    \includegraphics[width=0.9\columnwidth,height=0.4\columnwidth, keepaspectratio]{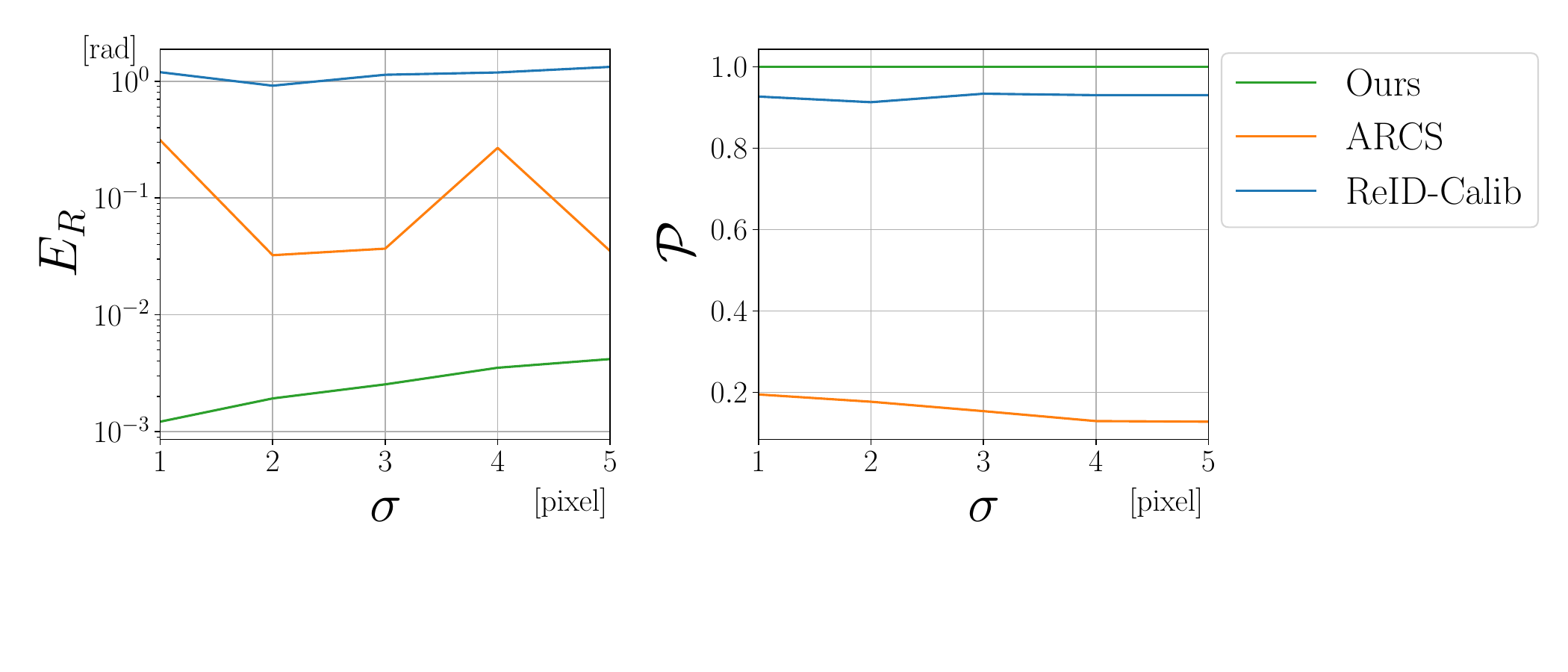} 
\end{tabular}
\vspace{-10mm}

\caption{Influence of noise on the accuracy of rotation and correspondence estimations. $\sigma$ represents the level of noise affecting joint positions.  }
\label{fig:exp_noise_level}
\end{figure}

\begin{table}[t]
    \centering
    \resizebox{\columnwidth}{!}{%
    \begin{tabular}{cccccccc}
    \toprule
& &    & $E_R$[\si{\radian]} & $E_\mathbf{t}$ & $E_\mathrm{2D}$[\si{px}]& $E_{\delta}$[\si{frame}]& $\mathcal{P} \uparrow$ \\
          \midrule
        \multirow{3}{*}{\shortstack{\textit{170915\_toddler5} \\ \lbrack 7,12,20,2\rbrack }}
&\multirow{3}{*}{w/o RANSAC}
& Min  &  0.001 &  0.000 &        1.681 &       0 &   0.976 \\
&& Max  &  0.005 &  0.007 &        1.745 &     0 &     0.983 \\
&& Mean &  0.002 &  0.003 &         1.709 &    0 &     0.979 \\
        \midrule
        \multirow{6}{*}{\shortstack{\textit{160422\_ultimatum1} \\ \lbrack 6,11,12,17\rbrack}}

&\multirow{3}{*}{w/o RANSAC}
&Min  &  0.001 &  0.001 &         1.713 &   0 &      0.002 \\
&&Max  &  2.273 &  1.263 &      103.557 &   5 &       0.983 \\
&&Mean &  0.678 &  0.364 &       32.125 &   2 &       0.607 \\
\cmidrule{2-8}
&\multirow{3}{*}{w/ RANSAC}

&Min  &  0.001 &  0.000 &         1.733 &      0 &   0.892 \\
&&Max  &  0.044 &  0.060 &        4.908 &     5 &     0.983 \\
&&Mean &  0.011 &  0.016 &         2.546 &   1 &      0.965 \\
        \midrule
        \multirow{6}{*}{\shortstack{\textit{160906\_pizza1} \\ \lbrack 2,12,21,28\rbrack }} 

&\multirow{3}{*}{w/o RANSAC}
&Min  &  0.001 &  0.001 &         1.686 &   0 &      0.018 \\
&&Max  &  2.761 &  1.369 &      147.295 &   5 &       0.983 \\
&&Mean &  0.245 &  0.134 &       16.177 &    0 &      0.888 \\
\cmidrule{2-8}
&\multirow{3}{*}{w/ RANSAC}
&Min  &  0.001 &  0.001 &        1.704 &     0 &     0.842 \\
&&Max  &  0.114 &  0.099 &       9.285 &     4 &      0.983 \\
&&Mean &  0.022 &  0.020 &        3.400 &     0 &     0.955 \\   
    \bottomrule
    \end{tabular}%
    }

    \caption{Unsynchronized pairwise registration.}

    \label{table:rigid_registration_sync}
\end{table}

\subsection{Robustness to Noise}
We use motion capture datasets to evaluate the performance in terms of establishing matches and estimating camera poses and temporal offset. We select three scenarios \st{\textit{170915\_toddler5}, \textit{160906\_pizza1}, and \textit{160422\_ultimatum1}} from CMU Panoptic dataset~\cite{Joo_2017_TPAMI} covering a wide variety of human activities, and used four cameras for evaluation. Additionally, we generate synthetic data by transforming ground truth 3D poses from the world coordinate system into each camera view coordinate system. We then inject Gaussian noise into the ground truth 3D points so that the standard deviation of their reprojection errors has $\sigma = 1, \dots, 5$ pixels. We also set the window size $T = 5$.
	
Fig.~\ref{fig:exp_noise_level} shows the average rotation errors $E_R$ on \textit{170915\_toddler5} at different noise levels. At each noise level, we synthesized 10 noisy observations, and calculated the average rotation errors between the four cameras.  In this evaluation, since ARCS and ReID-Calib do not estimate the temporal offset, all methods used the input with the ground truth temporal offset.  All methods also used the ground truth tracking per view.
For a fair comparison, our algorithm calculates the average of 10 random initial rotations for each trial. The results for the rotation error $E_{R}$ and precision $\mathcal{P}$ indicate that, even with noise simulating errors inherent in the 3D pose estimation process, the proposed method establishes reliable matches and estimates accurate rotations. It outperforms the baselines~\cite{yan2021wide-baseline,Peng2022ARCSAR} across all noise levels.
	
Table~\ref{table:rigid_registration_sync} shows the quantitative evaluation of spatiotemporal calibration and multi-person association in the three scenarios. For this experiment, we assumed a noisy environment with $\sigma=3$. \st{the relative time offset of $+5$ frames, the search range of temporal candidates of $+10$ frames. and random initial rotations.} Since \textit{160906\_pizza1} and \textit{160422\_ultimatum1} involve non-overlapping observations between views, we apply the outlier removal process using RANSAC (Sec.~\ref{sec:RANSAC}). The proposed method achieves accurate pairwise registration and temporal alignment simultaneously, regardless of persons’ activities and view configurations.

The processing time for \textit{170915\_toddler5} was approximately $38.11\si{s}$ and $192.98\si{s}$ with and without temporal alignment, respectively, on an Apple M1 Pro.

\begin{table}[!t]
    \centering
     \begin{tabular}{cc|cccc}\toprule
                & & 8-pts~\cite{zhang2000flexible} & PnP~\cite{Hartley00} & Inc. SfM~\cite{schonberger2016structure}& Ours \\ \midrule
                \multirow{3}{*}{$E_R$[\si{\radian}]} 
                & MI &0.001 &0.006  &0.037   & 0.347\\
                & BA &0.001 & 0.006 &0.038  & 0.072\\
                & IBA& -    & -     & -     & 0.006\\ \midrule
                \multirow{3}{*}{$E_\mathbf{t}$} 
                & MI &0.388 & 0.328 & 0.032 & 0.184  \\
                & BA &0.194 & 0.185 & 0.031 & 0.075 \\
                & IBA& -    & -     & -     & 0.004\\ \midrule
                 
                \multirow{3}{*}{$E_\mathrm{2D} [\si{px}]$} 
                & MI & 0.519& 1.699 & 7.804 & 93.824 \\
                & BA & 0.469& 1.416 & 4.689 & 12.665 \\ 
                & IBA& -    & -     &   -   & 3.778 \\ \bottomrule
                 
          \end{tabular}
    \caption{Comparison with conventional methods.}
    \label{table:comp_existing}
\end{table}

\subsection{Comparison with Chessboard-Based Calibration}

Table~\ref{table:comp_existing} shows calibration errors obtained from 100 trials in comparison with conventional methods using reference objects: the 8 point algorithm (8-pts)~\cite{zhang2000flexible}, the perspective-n-point (PnP)~\cite{Hartley00}, and the incremental SfM~\cite{schonberger2016structure}.  We randomly placed 10 multiple checkerboards in front of each pair of cameras. To simulate real-world conditions, we utilized the camera setting of \textit{soccer1\_6} and injected Gaussian noise with $\sigma = 0.5$ into the projected points in the images.  As Table~\ref{table:comp_existing} shows, the conventional methods achieve better reprojection error $E_\mathrm{2D}$.  Their errors in translation $E_\mathbf{t}$, however, are significantly high compared with ours.  This suggests that their calibration results can be considered as overfitting, while ours works robustly.

\section{Conclusion and Discussion}

We proposed a spatiotemporal multi-camera calibration method that estimates camera extrinsics, temporal offsets, and cross-view associations from multi-view videos, using persons in the scene as the calibration target.  Our method formulates the task of associating multiple people and calibrating multiple cameras as a single registration problem of two sets of 3D points on a unit sphere obtained by encoding 3D human poses of multiple people in motion. This registration is performed in a probabilistic manner, alternately updating spatiotemporal alignments and soft assignments for multi-person associations until geometric consistency is satisfied between views. The pairwise results from the multi-view setup are unified into a single nonlinear optimization framework that renders the spatiotemporal calibration into a single objective function by leveraging the linearity of 3D human motion in a short period of time.

Extensive experiments were conducted to validate the effectiveness and applicability of the proposed method. Our method demonstrated robust performance compared to the existing point-set registration and extrinsic multi-camera calibration using people. Moreover, we incorporated the temporal alignment into our framework and verified its accuracy and applicability for spatiotemporal multi-camera calibration. \\
We believe that our calibration method can serve as a practical tool in numerous real-world scenarios, eliminating the need for specialized equipment and meticulous operations when deploying multi-view systems.

\textbf{Limitation and Future work.} \textcolor{black}{Resolving global scale ambiguity remains challenging; however, this issue can be mitigated by incorporating scene priors. For example, in the \textit{170915\_toddler5} scenario, the length of a baseball bat held by a toddler can be used as a reference to ensure scale consistency. We validated that the translation error $E_\mathbf{t}$ was \SI{3.44}{cm}, with an average distance of \SI{1.93}{m} between paired cameras in the multi-camera network.} Since our method relies on appearance features, it may encounter errors in rotation estimation when calibration targets have similar poses (\eg a child with his/her mother mimicking the dance moves). However, the epipolar constraint applied during the translation estimation step mitigates this issue. We assumed that all cameras start recording almost simultaneously to limit the search space of the temporal offset.  In future work, we will explore integrating intrinsic calibration in our framework. \textcolor{black}{Our current implementation is designed for offline calibration, but improving computational efficiency is a key step toward real-time applications such as online calibration of dynamic cameras.}

\section*{Acknowledgement}

This work was in part supported by
JSPS %
20H05951 %
and 22H05654.

{
\bibliographystyle{IEEEtran}

\bibliography{egbib}
}
\end{document}